\documentclass{article}

\usepackage{arxiv}

\usepackage{adjustbox}

\usepackage{booktabs} 
\usepackage{subcaption}
\usepackage{amsmath}
\usepackage{amssymb}
\usepackage{mathtools}
\usepackage{amsthm}
\usepackage{multirow}
\usepackage{stfloats}
\usepackage{graphicx}
\usepackage{colortbl}
\usepackage{xcolor}
\usepackage{makecell}
\usepackage{amsmath}
\definecolor{Turquoise}{RGB}{64,224,208}

\usepackage[utf8]{inputenc} 
\usepackage[T1]{fontenc}    
\usepackage{hyperref}       
\usepackage{url}            
\usepackage{booktabs}       
\usepackage{amsfonts}       
\usepackage{nicefrac}       
\usepackage{microtype}      
\usepackage{lipsum}		
\usepackage{graphicx}
\usepackage[square,sort,comma,numbers]{natbib}
\usepackage{doi}

\title{Convolution-based Probability Gradient Loss for Semantic Segmentation}

\author{ {\hspace{1mm}Guohang Shan} \\
	Mogo Auto Intelligence and Telematics Information Technology Co., Ltd\\
	Beijing, China, 100013 \\
	\texttt{shanguohang@zhidaoauto.com} \\
	\And
	{\hspace{1mm}Shuangcheng Jia} \\
	Mogo Auto Intelligence and Telematics Information Technology Co., Ltd\\
Beijing, China, 100013\\
	\texttt{jiashuangcheng@zhidaoauto.com} \\
}

\renewcommand{\shorttitle}{\textit{arXiv} Template}

\hypersetup{
pdftitle={A template for the arxiv style},
pdfsubject={q-bio.NC, q-bio.QM},
pdfauthor={David S.~Hippocampus, Elias D.~Striatum},
pdfkeywords={First keyword, Second keyword, More},
}

\begin{document}
\maketitle

\begin{abstract}
In this paper, we introduce a novel Convolution-based Probability Gradient (CPG) loss for semantic segmentation. It employs convolution kernels similar to the Sobel operator, capable of computing the gradient of pixel intensity in an image. This enables the computation of gradients for both ground-truth and predicted category-wise probabilities. It enhances network performance by maximizing the similarity between these two probability gradients. Moreover, to specifically enhance accuracy near the object's boundary, we extract the object boundary based on the ground-truth probability gradient and exclusively apply the CPG loss to pixels belonging to boundaries. CPG loss proves to be highly convenient and effective. It establishes pixel relationships through convolution, calculating errors from a distinct dimension compared to pixel-wise loss functions such as cross-entropy loss. We conduct qualitative and quantitative analyses to evaluate the impact of the CPG loss on three well-established networks (DeepLabv3-Resnet50, HRNetV2-OCR, and LRASPP\_MobileNet\_V3\_Large) across three standard segmentation datasets (Cityscapes, COCO-Stuff, ADE20K). Our extensive experimental results consistently and significantly demonstrate that the CPG loss enhances the mean Intersection over Union.
\end{abstract}

\section{Introduction}
\label{sec:intro}
Semantic segmentation, a fundamental challenge in computer vision, involves assigning specific semantic category labels to individual pixels within an image. In recent years, considerable advancements have been made in this field, with powerful convolutional neural networks, including FCN\cite{Long2015CVPR}, DeepLab series networks\cite{chen2014semantic,chen2017deeplab,chen2017rethinking,Chen_2018_ECCV}, UNet\cite{ronneberger2015u}, HRNet\cite{wang2020deep}, BiSeNet\cite{Yu_2018_ECCV,yu2021bisenet}, SFNet-R18\cite{inbook2}, showcasing significant progress. Additionally, novel network structures and concepts, such as residual blocks\cite{He_2016_CVPR}, dilated convolution\cite{chen2017deeplab}, Denseblock\cite{Huang_2017_CVPR}, Transformer\cite{NIPS2017_3f5ee243}, multi-branch, and multi-scale\cite{Sun_2019_CVPR}, etc., have been developed. Despite these advancements, an emerging concern is the elevated false detection rate near object boundaries, as highlighted in prior studies\cite{inbook,Bertasius_2016_CVPR,Chen_2016_CVPR}. This issue is particularly pronounced in test cases involving slender and strip-shaped objects.
Many current networks exhibit substantial complexity and parameter quantities, suggesting untapped potential for improvement. With this in mind, our focus is on enhancing network performance by refining the training loss function.

Pixel-level loss functions, typically employed for network convergence, include Cross Entropy (CE) loss.
\begin{equation}
  L_{CE} = \frac{1}{N}\sum_{n=1}^{N}\sum_{c=1}^C(y_{n,c}log(p_{n,c})+(1-{y_{n,c}})log(1-p_{n,c}))
  \label{equation1}
\end{equation}

In this mathematical context, let $N$ denote the number of pixels, and $C$ represent the number of target types. The variables $y$ and $p$ signify the probabilities of a pixel belonging to a specific category, where 
$y\in$\{0, 1\} represents the true value. The predicted probability $p\in$[0, 1] is derived by applying the $sigmoid$ function to the network's output. This operation is commonly referred to as \textbf{BCELOSS} or \textbf{BCEWithLogitsLoss} in PyTorch\cite{web3}. Alternatively, the application of the $softmax$ function allows the omission of the segment following the ‘+,’ denoted as \textbf{CrossEntropyLoss} in PyTorch.

As indicated in Equation 1, the Cross Entropy (CE) loss is computed on a pixel-by-pixel basis. However, this loss function, by not considering the substantial interdependence between pixels\cite{1284395}, may constrain the overall performance of networks.

Currently, various models have been proposed to establish relationships between pixels, including Conditional Random Field (CRF)-based methods\cite{NIPS2011_beda24c1,Shen_2017_CVPR,8278309}, pixel affinity-based methods\cite{Ke_2018_ECCV,Ahn_2018_CVPR,Xu_2021_ICCV}, mutual information-based methods\cite{NEURIPS2019_a67c8c9a}, and boundary-based methods\cite{wang2022active}. Our approach, however, adopts a novel perspective by building pixel relationships through gradients.

In line with \cite{inbook}, our work is similarly motivated by the challenge posed by "most existing state-of-the-art segmentation models struggling to handle error predictions along the boundary." In \cite{inbook}, the author provides a detailed description of the specific phenomenon associated with this issue, ultimately contributing to a poor segmentation effect for slender or small-area objects in the network.

We examined the characteristics near the boundary in selected semantic segmentation results and observed that the probability did not exhibit a sharp enough change, as illustrated in Figures 2, 5, and 6. It is typical for two or more categories to possess similar predictive probabilities at these pixels, and a slight advantage for one category may determine the prediction result. Our objective is to enhance the category-probability advantage on pixels by amplifying the probability difference at adjacent pixels near the boundary. This enhancement can be realized by increasing the probability gradient along pixels, in other words.

Building upon the aforementioned concept, this paper introduces the CPG loss. Sobel-like operators are employed to conduct convolution on both ground-truth and predicted probabilities to get their gradients. The ground-truth gradient is utilized to derive the boundary for each category in the image, so the focus of the loss function can be directed towards pixels along the boundary. Ultimately, Mean Square Error (MSE) loss is employed to compute the loss value, as illustrated in Equation 2, where $g$ represents the probability gradient, $N_c^+$ denotes the number of edge pixels for the $c$-th category, and $N^+$ indicates the total pixel count along the boundaries.
\begin{equation}
  L_{CPG} = \frac{1}{N^+}\sum_{c=1}^{C}\sum_{n=1}^{N_c^+}(g_{n,c}^{gt}-g_{n,c}^{pred})
  \label{equation2}
\end{equation}
On one hand, unlike CE loss, which only constrains the value of a single pixel, CPG loss establishes relationships between a pixel and its surrounding pixels through convolution. On the other hand, owing to the presence of the ground-truth probability gradient, CPG loss serves as an end-to-end loss function in practice. This implies it has a destination (0.0), distinguishing it from many similar losses such as Boundary loss\cite{pmlr-v102-kervadec19a} and Region Mutual Information (RMI) loss.

We evaluated the efficacy of the CPG loss using three well-established networks: DeepLabv3-Resnet50\cite{chen2017rethinking}, HRNetV2-OCR\cite{DBLP-journals/corr/abs-1909-11065}, and LRASPP\_MobileNet\_V3\_Large\cite{Howard_2019_ICCV}, across three standard datasets—Cityscapes\cite{Cordts_2016_CVPR}, COCO-Stuff\cite{Caesar_2018_CVPR}, ADE20K\cite{Zhou_2017_CVPR,zhou2019semantic}. The results demonstrate that CPG loss significantly enhances the mean Intersection over Union (mIoU) for all three networks. Furthermore, our experimentation with RMI loss indicates that CPG loss can collaboratively enhance network performance when employed in conjunction with RMI loss.

Main contributions of this paper:

1. We introduce the Convolution-based Probability Gradient (CPG) loss, which effectively enhances the performance of semantic segmentation networks when used in combination with CE loss.

2. This loss function is network-agnostic, allowing seamless integration into most existing semantic segmentation networks without the need for additional modifications.

3. Extensive comparative experiments have been conducted, and the results have been analyzed to demonstrate the effectiveness of the proposed loss function and provide insights into optimal utilization across different networks.

Relevant codes for implementing the CPG loss are available.
[https://github.com/MoriartyShan/Convolution-based-Probability-Gradient-Loss.git]

\section{Related Works}
\label{sec:formatting}
\subsection{The Sobel Operator}

The Sobel operator\cite{gonzalez2014improved,Web_1} finds application in image processing and computer vision, particularly in edge detection algorithms\cite{4767851,chetia2021quantum,gonzalez2016improved}. Functioning as a discrete differentiation operator, it computes pixel intensity gradients in an image using convolution. It is formally defined in Equation 3.

\begin{equation}
  K_x = \begin{bmatrix}
  -1 & 0&1\\
-2 &0&2\\
-1&0&1
\end{bmatrix}
  \label{equation3}
\end{equation}
\begin{equation}
 K_y = K_x^T
  \label{equation4}
\end{equation}

$K_x$ and $K_y$ are applied in the horizontal and vertical directions, respectively. In our study, both $K_x$ and $K_y$ are multiplied by 0.5, ensuring that the maximum element value in the kernel becomes 1.0.

For a pixel in the image, located at the $i$-th row and $j$-th column, its gradient can be calculated as follows:

\begin{equation}
  G_x(i,j)=\sum_{r=0}^{M-1}\sum_{c=0}^{M-1}K_x(r, c)\times I(i+r-\frac{M-1}{2}, j+c-\frac{M-1}{2})
  \label{equation5}
\end{equation}

Where $I(i, j)$ represents the intensity of the pixel in the image, and $M$ denotes the size of the convolution kernel, with $M$ set to 3 in this example. Upon replacing $K_x$ with $K_y$, the resulting gradient is denoted as $G_y$.

As the x-coordinate and y-coordinate directions are orthogonal, the magnitude and direction of the gradient can be obtained as Equations 6 and 7, respectively.
\begin{equation}
  G(i,j)=\sqrt{G_x^2(i,j)+G_y^2(i,j)}
  \label{equation6}
\end{equation}
\begin{equation}
  \theta(i,j)=atan2(G_y(i,j),G_x(i,j))
  \label{equation7}
\end{equation}
and $atan2$ is a 2-argument arctangent function, which can get the input 2-D vector’s angle within $(-\pi, \pi]$.

The calculation for horizontal and vertical gradients is independent, and the convolution kernels are relatively small, resulting in low overhead. The gradient magnitude indicates the presence of an edge at a given position, and the Sobel operator excels at edge detection. However, the edges computed using this method are coarse, posing a significant challenge for images with high-frequency variation or noise. Typically, additional methods are required to enhance the accuracy of edge detection\cite{4767851}. Figure 1 illustrates the gradient result computed by the Sobel operator.

\begin{figure}
  \centering
  \begin{subfigure}{0.45\linewidth}
    \includegraphics[width=1\linewidth]{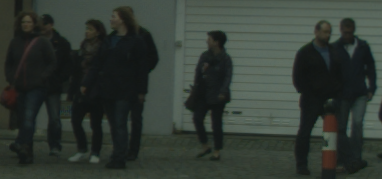}
    \caption{Input image.}
  \end{subfigure}
  \hfill
  \begin{subfigure}{0.45\linewidth}
    \includegraphics[width=1\linewidth]{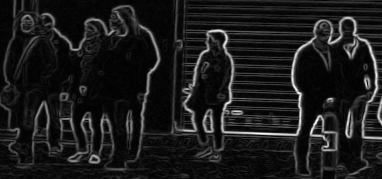}
    \caption{image gradient with Sobel.}
  \end{subfigure}
 
  \begin{subfigure}{0.45\linewidth}
    \includegraphics[width=1\linewidth]{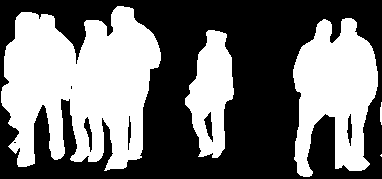}
    \caption{Ground truth of person.}
    \label{fig:short-a}
  \end{subfigure}
  \hfill
  \begin{subfigure}{0.45\linewidth}
    \includegraphics[width=1\linewidth]{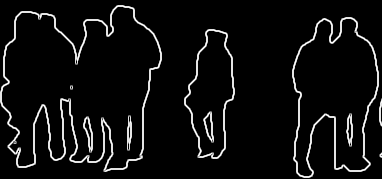}
    \caption{GT gradient with Sobel.}
  \end{subfigure}
  \caption{Results from applying the Sobel operator to process both the input image and ground truth (GT). The outcomes are normalized, with brighter pixels approaching 1.0 and darker pixels approaching 0.0. Convolution of the original image with the Sobel operator detects edge, with the non-zero gradient magnitude indicating edges. However, noise in the image introduces false detections. Conversely, convolving the ground truth with the Sobel operator allows for nearly error-free detection of category boundaries.}
  \label{figure1}
\end{figure}
Moreover, there exist operators of varying sizes akin to the Sobel operator, which can be employed to compute pixel-intensity gradients in an image. The discussion, as referred to on $StackOverflow$ \cite{web2,lateef2008expansion}, indicates that the convolution kernel used for calculating pixel gradients can be generated using Equations 8 and 9:
\begin{equation}
  K_x^{2m+1}(i,j)=\begin{cases}
  0 & j = m\\
  \frac{j-m}{(i-m)^2+(j-m)^2} & otherwise
  \end{cases}
  \label{equation8}
\end{equation}
\begin{equation}
  K_y^{2m+1}=K_x^{2m+1^T}
  \label{equation9}
\end{equation}

Here $2m+1=M$ represents the size of the convolution kernel, with $(i, j)$ denoting the index of the element in the operator.

In our study, the Sobel operator computes gradients for both Ground Truth (GT) and predictions probability. GT comprises organized binary information without noise, with pixel intensity changes occurring exclusively at object edges. The Sobel operator yields non-zero gradients at the edges, while gradients for inner/outer objects are set to 0. As illustrated in Figure 1(c) and (d), the edge is accurately obtained based on gradient magnitude.

In addition to the Sobel operator, we utilize convolution operators of various sizes (primarily 5$\times$5 and 7$\times$7), generated by Equations 8 and 9, to obtain pixel intensity gradients in the image. And larger convolution kernels tend to result in broader boundaries. Throughout this paper, CPG$^M$ denotes the use of a convolution kernel with a size of $M{\times}M$ in the calculation of CPG loss.

\section{Methodology}

\subsection{Motivation for Probability Gradient}

Figure 2 depicts the ground-truth and predicted probability along a continuous segment of pixels in the output of a binary classification network. In semantic segmentation, a pixel belongs to a particular category, while its closely adjacent pixels may not. As indicated by the green curve, the 5th pixel has a value of 1, while the 4th pixel has a value of 0. Due to convolutional characteristics, the predicted results of convolutional networks typically exhibit gradual changes, unlike the more abrupt changes in ground truth (GT), as illustrated by the red curve.

\begin{figure}[t] 
	\centering
		\begin{minipage}[t]{0.48\linewidth}
			\centering
			\includegraphics[width=0.8\linewidth]{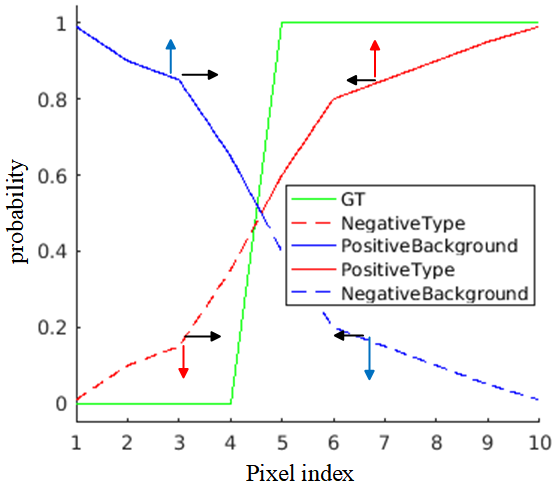}

   \caption{Predicted probability for a continuous segment of pixels near the object edge in binary classification results. The red curve represents the network's predicted probability for the target category, the blue curve for the background category, and the green curve for the ground truth of the target category. Solid lines indicate pixels belonging to the respective category, while dotted lines indicate otherwise.}
   \label{figure2}
  
		\end{minipage}%
   \hspace{10pt}
		\begin{minipage}[t]{0.48\linewidth}
  \centering
   \includegraphics[width=0.9\linewidth]{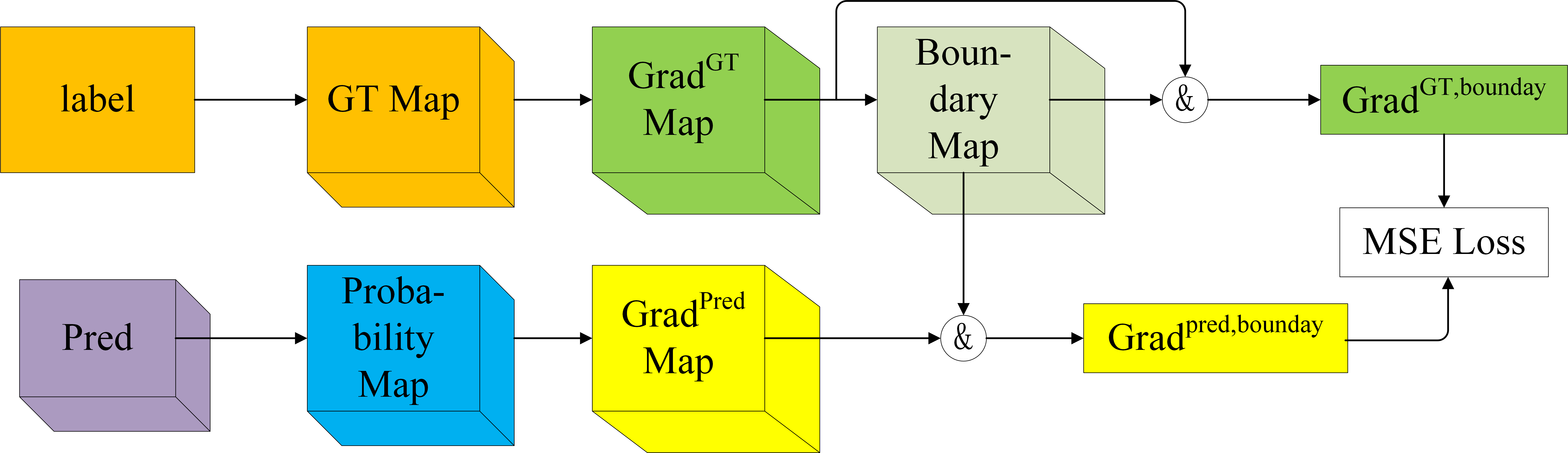}
   \caption{Illustration of the CPG loss pipeline. $Grad^{GT}$ and $Grad^{Pred}$ Maps are generated using the same method, and the Boundary Map is derived from $Grad^{GT}$.}
   \label{figure3}
		\end{minipage}%
\end{figure}




The pixel category is determined by selecting the maximum probability among all classes. However, when the predictive probability along a pixel curve changes gradually near the boundary, with minimal differences between categories, there is a risk of a slight advantage for one category, potentially leading to errors.

CE loss exclusively considers the pixel value and overlooks the numerical relationship with its neighbors. Consequently, during training, CE loss tends to align predicted values with true values (0 or 1), optimizing along the directions indicated by the red and blue arrows in Figure 2. However, it fails to ensure the predicted results exhibit a step characteristic at the edge position.

Recognizing the effectiveness of gradients in describing step characteristics, we considered comparing the prediction's gradient with its ground truth and using the error of the probability gradient as a loss. A large gradient not only widens the probability gap between categories but also, owing to its dependence on the pixel's neighborhood intensity, establishes pixels relationship. The optimization tendency of such a loss aligns both horizontally (as indicated by the black arrows in Figure 2) and vertically, bringing the probability curve closer to the ground truth.

Building upon these concepts, we introduce the Convolution-based Probability Gradient (CPG) loss.

\subsection{Implementation}
Figure 3 illustrates the CPG loss pipeline, while Figure 4 demonstrates the calculation of CPG loss for a specified image region.

\begin{figure}
  \centering
   \includegraphics[width=1.0\linewidth]{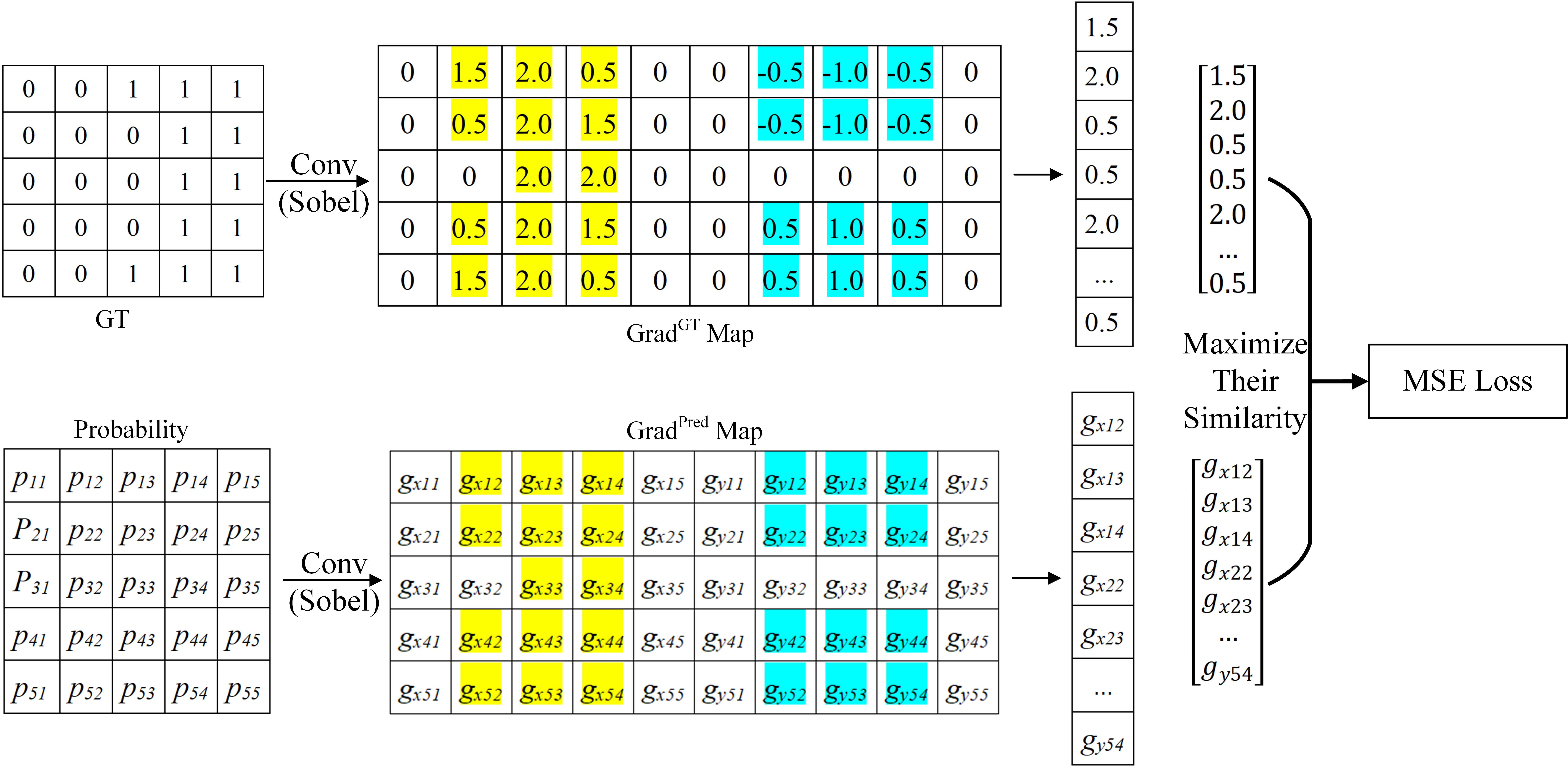}
   \caption{An image region and its corresponding CPG loss result. Following the same strategy, the CPG loss for the entire image can be calculated. The top row outlines the routine for processing ground truth, while the bottom row pertains to predicted results. CPG loss performs analogous operations on both $GT$ and $Pred$. First, solve $Grad$ Maps, then derive the boundary based on $Grad^{GT}$ Map, and finally extract $Grad$ Maps at the boundary while maximizing similarity between two $Grad^{Boundary}$ Maps. The gradient calculated is twice as large as the pixel count of the original image, as it encompasses gradient results in both x-coordinate and y-coordinate directions.}
   \label{figure4}
\end{figure}

\subsubsection{Generating Probability}

The input label typically shares the same size as the original image, with each pixel's value corresponding to the index of a category. Utilizing the \verb|one-hot| encoding method, we process the input label to obtain the ground-truth probability of each category for each pixel, creating a map referred to as $GT$ Map. If the total number of categories is $C$ and the pixel's value is $i$, a vector with $C$ elements is generated, where the $i$-th position holds the value 1 and others are set to 0. \verb|one-hot| encoding is a common method, and the formulas are as follows:
\begin{equation}
  \verb|one-hot|(i, C)=[l_0, l_1, l_2, \cdots, l_k, \cdots, l_{C-1}]
  \label{equation10}
\end{equation}
\begin{equation}
  l_k=\begin{cases}
  1, \quad if \quad k=i\\
  0,\quad otherwise
  \end{cases}
  \label{equation11}
\end{equation}

Elements in this vector represent the probability of the pixel belonging to each category. The $i$-th position having a value of 1 indicates that the pixel is 100\% associated with the $i$-th category, with the probability of other categories set to 0.

The output of the network ($pred$) requires processing through the $softmax$ function. This step allows us to predict the probability that the corresponding pixel belongs to each category, denoted as $probabilityMap$.

\begin{equation}
  probabilityMap(x, i)=\frac{e^{pred(x, i)}}{\sum_{j=0}^{C-1}e^{pred(x,j)}}
  \label{equation12}
\end{equation}
In which $x$ is the index of the pixel, $i$ represents the channel index of the pixel position, and $pred$ is the output of the network.

\subsubsection{Pixel Probability Gradient and Boundary}
Utilize Sobel-like operators for image convolution (Equation 5) on $GT$ Map to obtain the ground-truth pixel gradient, recorded as the $Grad^{GT}$ Map. It's important to note that the same Sobel operators, encompassing both x-coordinate and y-coordinate directions, are applied to each channel. $Grad^{GT}$ Map contains two directional gradients, resulting in a shape of $[C, 2, H, W]$, where $H$ and $W$ represent the height and width of the image, respectively.

As the internal pixels of objects in the $GT$ Map have a value of 1 and external pixels have a value of 0, the $Grad^{GT}$ Map value at these pixels is 0, as depicted in Figure 1(d). However, at the boundary pixels, owing to different values in the 8 (or 24, 48, depending on the size of the convolution kernel used) neighborhood, the $Grad^{GT}$ Map contains non-zero values. Utilizing $Grad^{GT}$ Map, it becomes straightforward to identify boundaries in the image, referred to as $BoundaryMap$. The shape of $BoundaryMap$ aligns with $Grad^{GT}$ Map because boundaries of each category are needed. Different-sized convolution kernels yield varying widths of boundaries.
\begin{equation}
   BoundaryMap(i)=\begin{cases}
  0, \quad Grad^{GT}(i)=0\\
  1,\quad otherwise
  \end{cases}
  \label{equation13}
\end{equation}

It should be noted that all calculations for the ground truth do not require backpropagation during training, so the related gradients during this process do not need to be retained, reducing memory overhead. Moreover, pre-calculating $GT$ Map and $BoundaryMap$ during data loading can save time in the training process.

Similar to $GT$ Map, the Sobel-like operator is applied to $probabilityMap$ (Equation 5) to obtain the gradient of the prediction result, denoted as $Grad^{pred}$.

\subsubsection{Gradient Similarity Evaluation}
$BoundaryMap$ is utilized to extract values at object boundary positions from $Grad^{GT}$ and $Grad^{pred}$. This focus on object borders is crucial as inaccurate classification often occurs in these regions. The objective is to evaluate the error of the probability gradient specifically at the boundary of an object rather than over the entire image. The gradients of truth and prediction at the object boundary are denoted as $Grad^{GT,Boundary}$ and $Grad^{pred,Boundary}$.
\begin{equation}
  Grad^{Boundary}(i)=Grad(i)\times BoundaryMap(i)
  \label{equation14}
\end{equation}

$Grad^{GT, Boundary}$ and $Grad^{pred, Boundary}$ capture true and predicted probability gradients in the x-coordinate and y-coordinate for each pixel at category boundaries.  Mean Square Error (MSE) loss is then applied to gauge their similarity, with $N^+$ denoting positions where $BoundaryMap$ is 1 in Equation 2.

The form of the final loss function is:
\begin{equation}
  L_{final}=L_{CE}+\alpha L_{CPG}
  \label{equation15}
\end{equation}

where $\alpha$ is the weight of CPG loss.

\section{Experiments}

We assess the performance of CPG loss on three standard datasets — ADE20K, Cityscapes, and COCO-Stuff — employing three different network types: DeepLabv3-Resnet50, HRNetV2-OCR, and LRASPP\_MobileNet\_V3\_Large.

It's worth noting that certain test results in this chapter may exhibit slight variations from those in the original paper. This discrepancy arises because we opted not to use the original paper's test parameter settings for each network. Instead, we followed our procedure to train and validate these models in this paper. The decision to deviate from the original settings is based on the complexity of some training steps in the original paper and the variations in test datasets and training parameters among different networks. Utilizing a consistent method for re-training these models not only simplifies the experimental process but also eliminates unnecessary interference.

\subsection{Datasets}
\noindent
\textbf {Cityscapes}\cite{Cordts_2016_CVPR,Cordts2015Cvprw} primarily focuses on urban road scenes with high-resolution images. We evaluate the semantic segmentation results using 19 out of the total 30 categories. The training set comprises 2975 images, and the validation set contains 500 images. Additionally, the dataset provides a substantial amount of coarse-labeled data. Inspired by \cite{DBLP:journals/corr/abs-2005-10821,Xie_2020_CVPR}, we leverage autolabeled data from Andrew Tao\cite{DBLP:journals/corr/abs-2005-10821}, which includes approximately 20K fine-labeled samples generated automatically based on coarse annotations.

\noindent
\textbf {ADE20K}\cite{Zhou_2017_CVPR,zhou2019semantic} was employed in the Imagenet Scene Parsing Challenge 2016 \cite{ILSVRC15}. This dataset offers rich and diverse object types and scenes, encompassing 150 segmentation categories. The training set comprises 20,210 images, and the validation set contains 2000 images.

\noindent
\textbf {COCO-Stuff}\cite{Caesar_2018_CVPR} is a panoramic segmentation dataset featuring 171 semantic segmentation categories. The training and test sets we utilize consist of 9K and 1K images, respectively.

\subsection{Networks}
In selecting test networks, our aim is to encompass various types and characteristics of excellent networks, rather than focusing on the latest and best-performing ones. This approach is designed to demonstrate the effectiveness of CPG loss across different network architectures. We have chosen the following three networks for evaluation:

\noindent
\textbf{HRNetv2-OCR}\cite{DBLP-journals/corr/abs-1909-11065}: HRNetV2-OCR, with 70.4M parameters, achieved top mIoU on the Cityscapes test set. Utilizing HRNetV2-W48 as backbone, it incorporates the object-contextual representations module to enhance semantic segmentation accuracy. The network also integrates advanced components such as Attention mechanism, Transformer\cite{NIPS2017_3f5ee243}, and multi-scale techniques. Many recent works build upon this foundation\cite{Borse_2021_CVPR,DBLP:journals/corr/abs-2005-10821,rs13214237}. For our experiments, we use the pre-trained weights provided by the author\cite{Sun_2019_CVPR,wang2020deep}, applying CPG loss only to the final output.

\noindent
\textbf{DeepLabv3-Resnet50}: DeepLabV3, with a ResNet-50 backbone and 42.3M parameters, employs dilated convolution and ASPP structure. These features are shared with many excellent networks\cite{Yang_2018_CVPR,Mehta_2018_ECCV,Chidanand_2021_WACV}. In our training, we use pre-trained weights from PyTorch\cite{web3}.

\noindent
\textbf{LRASPP\_MobileNet\_V3\_Large}\cite{Howard_2019_ICCV}: This Lite R-ASPP network model, with a MobileNetV3-Large backbone and 3.2M parameters, is a compact network designed for fast inference. It employs various simplification methods for computation efficiency. Pre-training weights from PyTorch\cite{web3} are used in our experiments.

\begin{figure}
  \centering
   \includegraphics[width=0.9\linewidth]{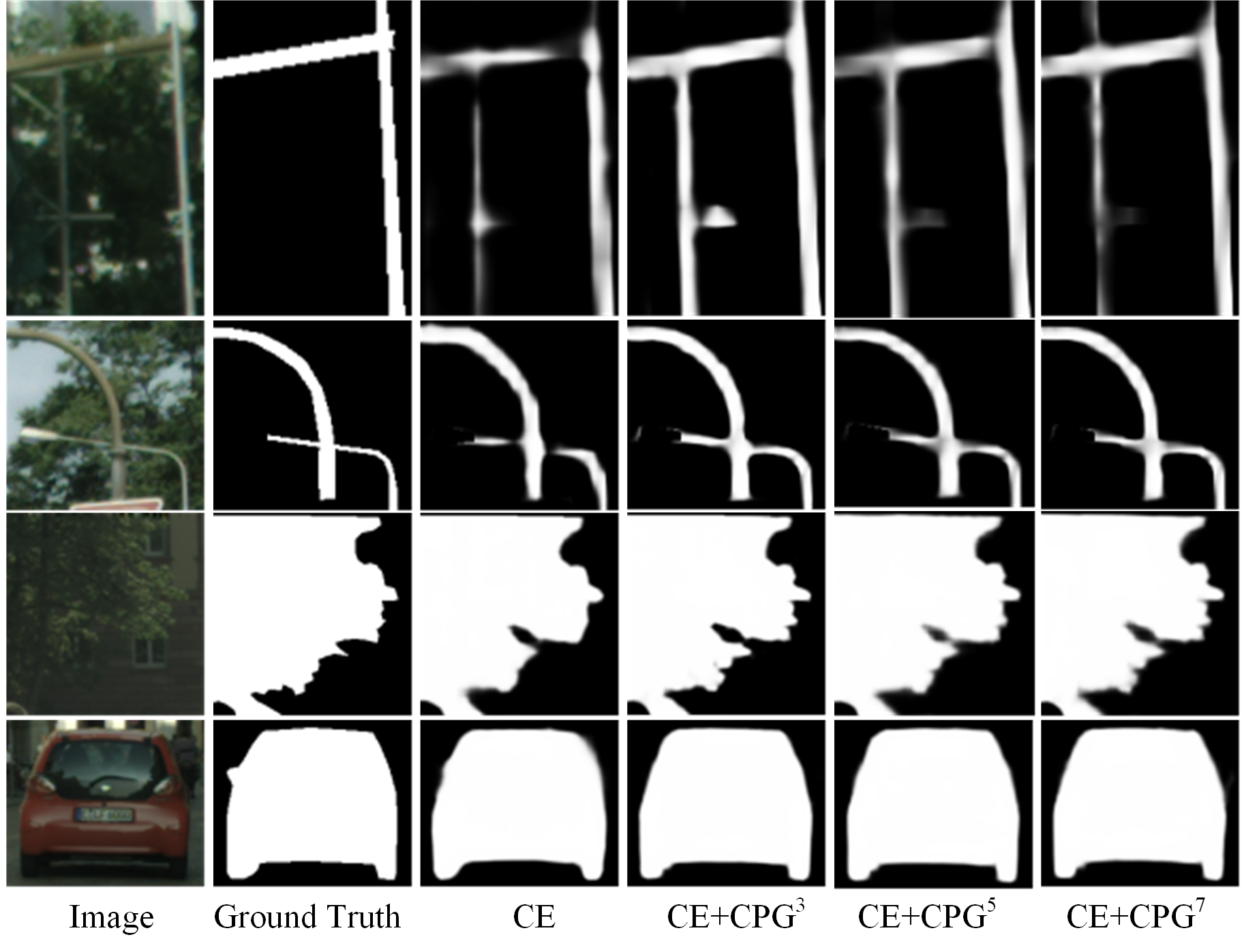}
   \caption{The predictive probability from DeepLabv3-Resnet50 on Cityscapes valset. The figure represents the likelihood of a pixel belonging to the target category. Brighter pixels indicate a closer probability to 1.0, while darker pixels signify proximity to 0.0. The first and second rows correspond to the category "pole," achieving a mIoU of 63.71\% with CE loss on the validation set. Combining CE loss with CPG$^7$ loss increases it to 70.23\%. In the third row, the category is "vegetation," yielding a mIoU of 92.23\% with CE loss, which improves to 93.07\% with CPG$^7$ loss. The fourth row represents the category "car," achieving a mIoU of 94.78\% with CE loss and 95.63\% with CPG$^7$ loss. Further details are provided in Table 1, indicating that the use of CPG loss enhances edge detection, reduces blurring, and reveals more detailed boundaries. In the given image, a vertical pole is situated in the middle of the first row. Although labeled as "other" in the Cityscapes dataset, which is not part of the training categories, it demonstrates strong activation within the "pole" category due to its similarity.}
   \label{figure5}
\end{figure}

\begin{figure}
  \centering
  \begin{subfigure}{0.48\linewidth}
    \includegraphics[width=1\linewidth]{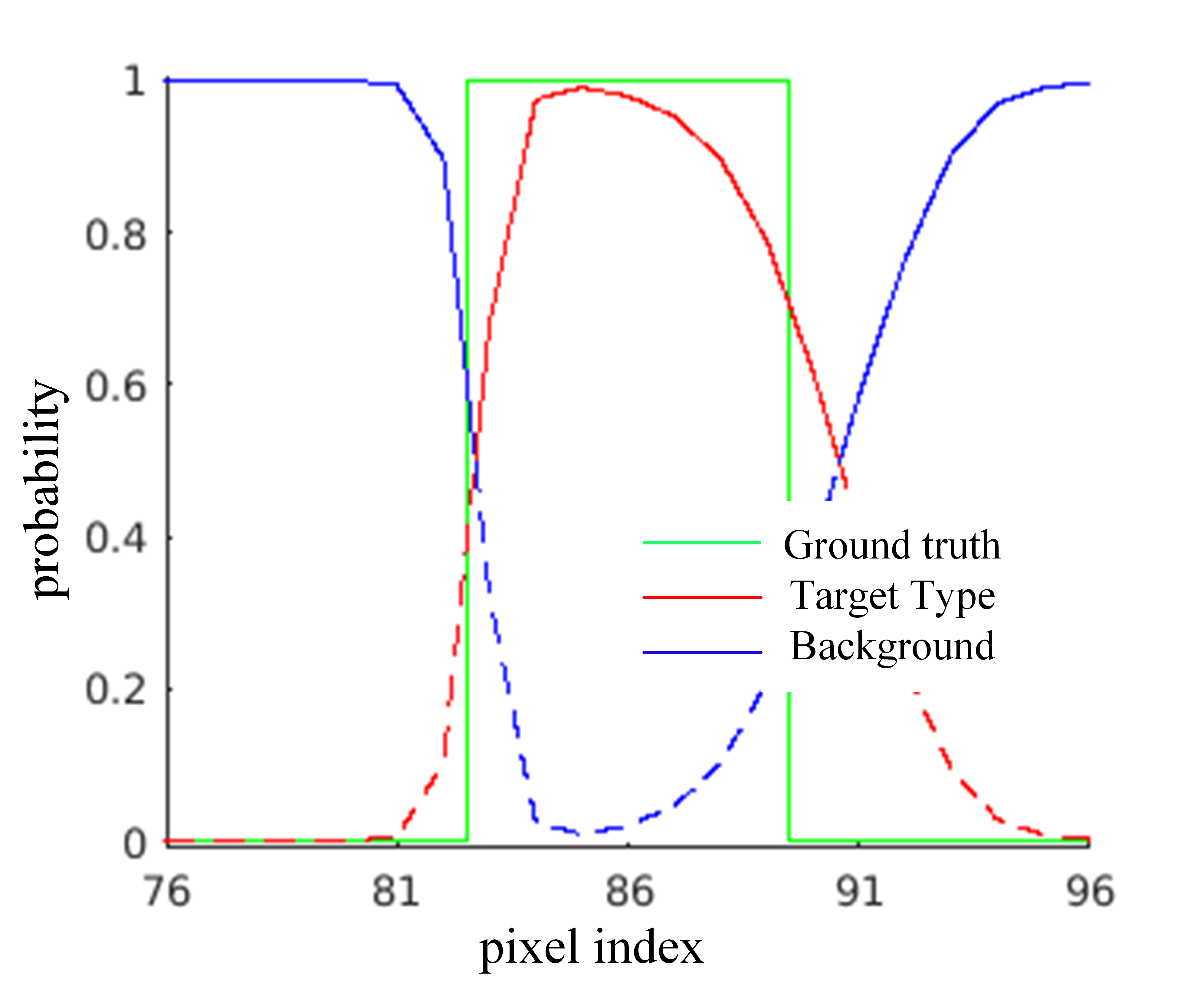}
    \caption{Probability of pixels in 78th row trained with CE loss.}
  \end{subfigure}
  \hfill
  \begin{subfigure}{0.48\linewidth}
    \includegraphics[width=1\linewidth]{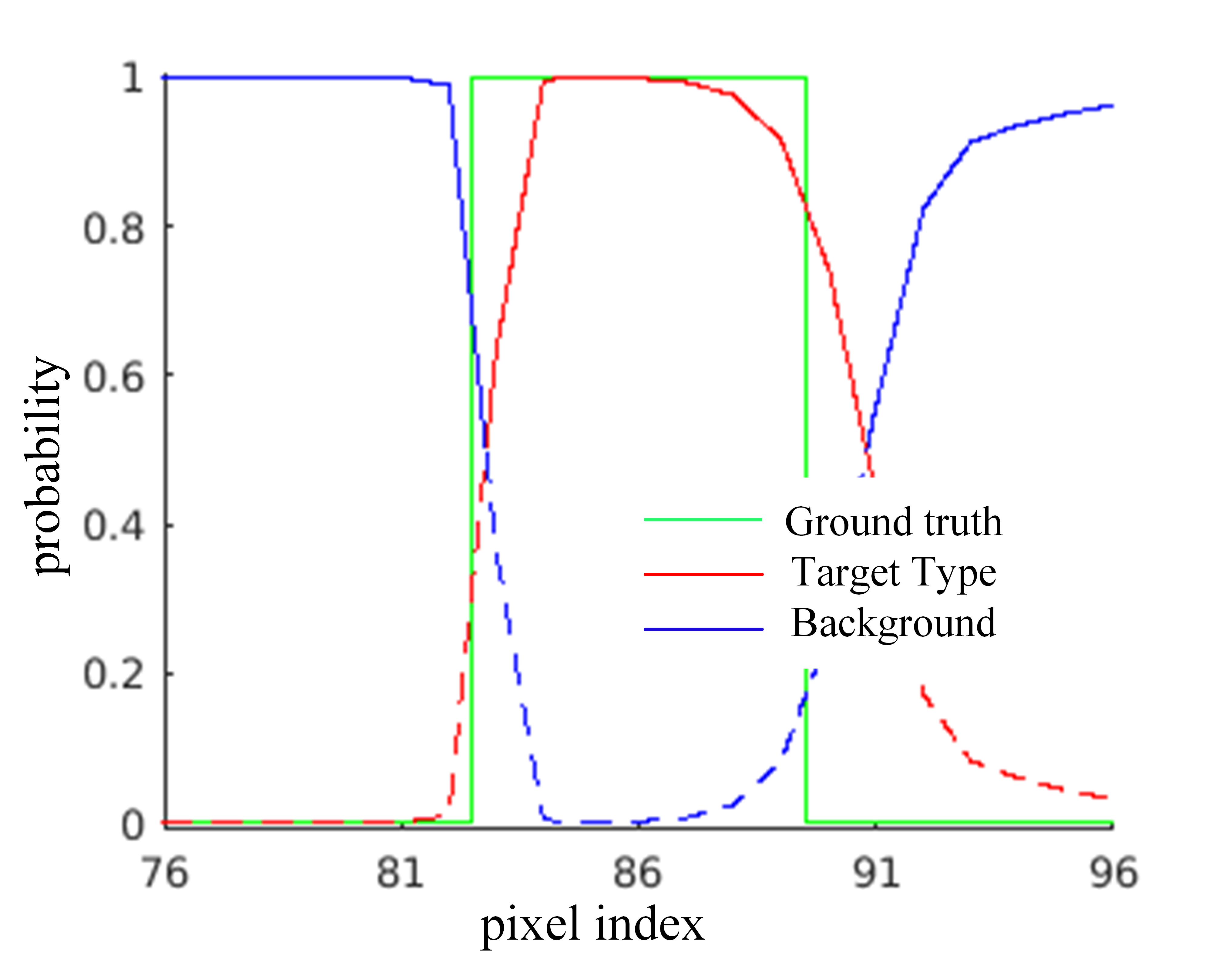}
    \caption{Probability of pixels in 78th row trained with CE+CPG$^3$ loss.}
    \label{fig:short-b}
  \end{subfigure}
 
  \begin{subfigure}{0.48\linewidth}
    \includegraphics[width=1\linewidth]{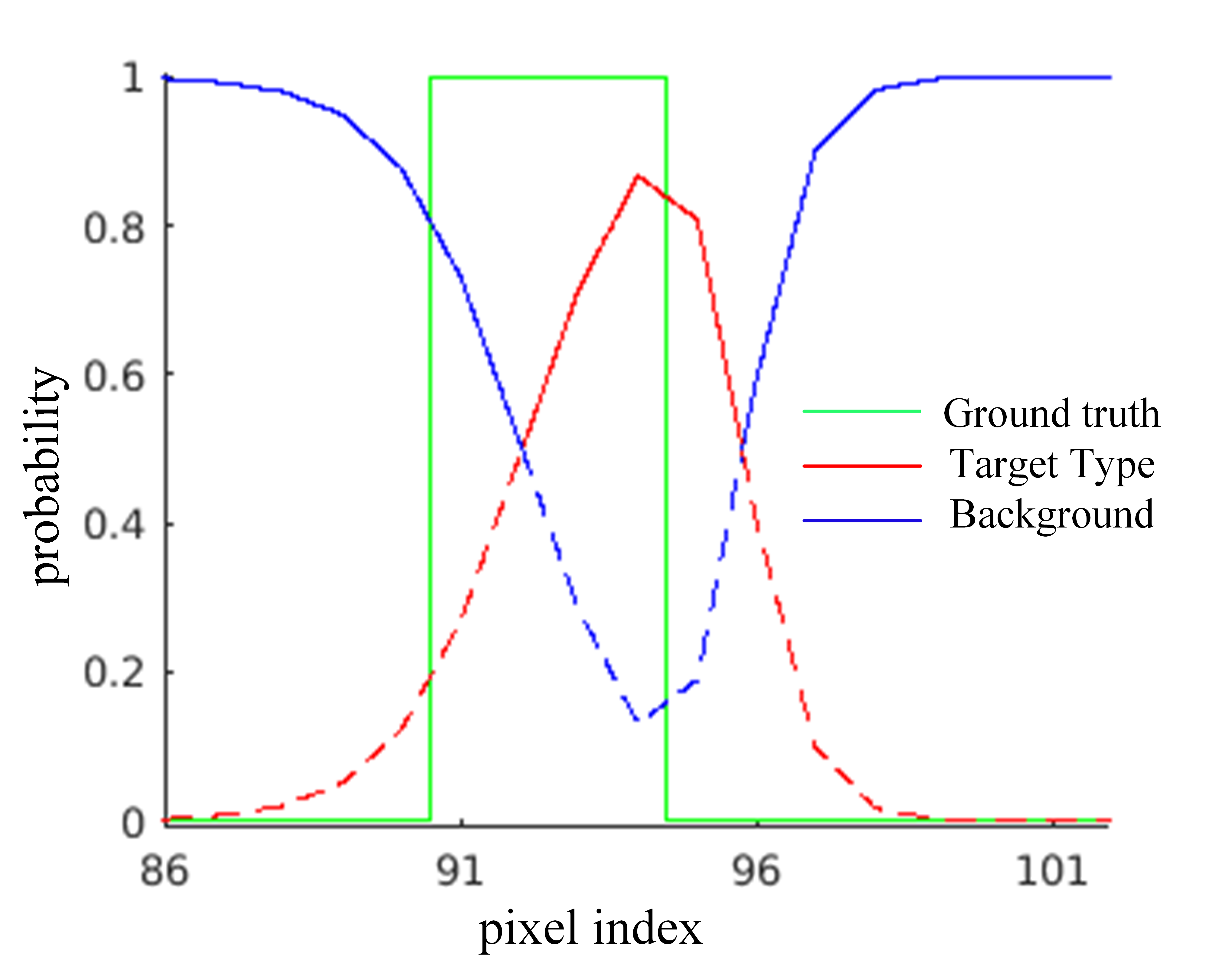}
    \caption{Probability of pixels in 104th column trained with CE loss.}
    \label{fig:short-c}
  \end{subfigure}
  \hfill
  \begin{subfigure}{0.48\linewidth}
    \includegraphics[width=1\linewidth]{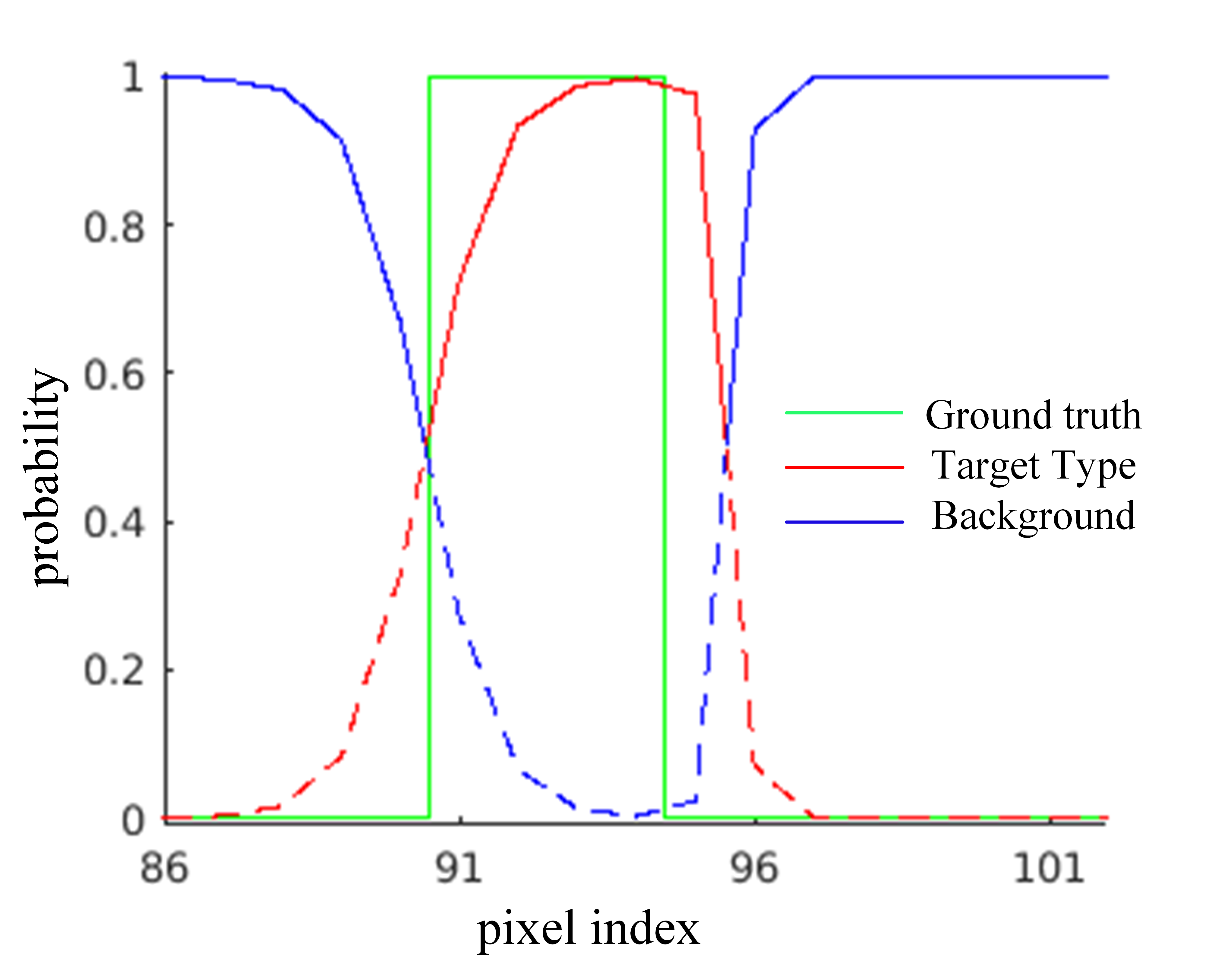}
    \caption{Probability of pixels in 104th col. trained with CE+CPG$^3$ loss.}
    \label{fig:short-d}
  \end{subfigure}
  \caption{Comparison of the result with/without CPG$^3$ loss. The red curve represents the network's predicted probability for the target category, the blue curve for the background category, and the green curve for the ground truth of the target category. Panels (a) and (b) correspond to images in the 1st row of Figure 5, while (c) and (d) correspond to images in the 2nd row of Figure 5. It is evident that the numerical values in (b) and (d) exhibit more rapid variations than those in (a) and (c), with peak and valley values closer to 1 and 0.}
  \label{figure6}
\end{figure}

\subsection{Training Details}
Following various tests, we opt for distinct training parameters tailored to each network and dataset to attain superior segmentation results. Specific details are outlined below.

\noindent
\textbf{Test environment}. Nvidia A100-SXM4-40GB, cuda 11.1, pytorch 1.8.0.

\noindent
\textbf{Learning rate policy}. The learning rate is denoted as $lr$, and the weight decay of the stochastic gradient descent optimizer is set to 1e-4. The learning rate undergoes decay based on the training epochs, employing the polynomial decay strategy. The calculation equation for the learning rate in an epoch is $lr \times (1-epoch/max\_epoch)^ {power}$, with $power$ set to 2.0.
\noindent
\textbf{Loss function}: Use \textbf{BCEWithLogitsLoss} as main loss on Cityscapes and \textbf{CrossEntropyLoss} on ADE20K and COCO-Stuff.

\noindent
\textbf{Data augmentation}:Some data augmentation methods are used for the training data, including random scaling of the input image with a factor of 0.5-2.0, and random left and right flips. No data augmentation is performed on test sets.

\noindent
\textbf{Cityscapes}: 
We utilize a training set comprising 2975 images for training and a validation set with 500 images for testing. The crop size is set to 1024$\times$2048. The batch sizes for DeepLabv3-Resnet50, HRNetV2-OCR, and LRASPP\_MobileNet\_V3\_Large are 12, 8, and 16, respectively. Additionally, the details of the auto-labeled set are as follows: in the 0 to ${(x-1)}$ epochs, we randomly select 0.5$\times$2975 images from both the training set and auto-labeled set to create a mixed set, ensuring the proportion of each category matches that of the training set. In the $x$ to last epoch, we use only the training set, where $x$=175 for LRASPP\_MobileNet\_V3\_Large and $x$=150 for the other two networks.

\noindent
\textbf{ADE20K} and \textbf{COCO-Stuff}: 
We use both the training set and the validation(test) set. Due to variations in the original picture sizes, many of which are small, we standardize the crop size to 520$\times$520. The batch sizes for DeepLabv3-Resnet50, HRNetV2-OCR, and LRASPP\_MobileNet\_V3\_Large are 16, 16, and 32, respectively.

\noindent
\textbf{Other details}. For the test case HRNetv2-OCR on COCO-Stuff, we set $max\_epoch$ = 110 and $lr$ = 0.001, and $max\_epoch$ = 200 and $lr$ = 0.014 for LRASPP\_MobileNet\_V3\_Large on Cityscapes. Other cases are $max\_epoch$ = 175 and $lr$ = 0.01.

\begin{table*}
  \centering
  \resizebox{\linewidth}{!}{
  \begin{tabular}{@{}lccccccccccccccccccc@{}}
    \toprule
    &road	&Side-walk	&building&	wall	&fence	&pole	&Traffic 
light&	Traffic
 sign&	Veget-
ation&	terrain&	sky&	person&	rider&	car&	truck&	bus	&train&	Moto-
rcycle&	bicycle \\

    \midrule
    CE&	98.19	&85.56	&92.40	&\colorbox{green}{58.36}	&\colorbox{green}{62.35}	&63.71	&66.92	&76.81&92.23&62.94&94.31&80.24&60.23&94.78&68.94&84.34&77.63&62.67&\colorbox{green}{75.64}\\
    
    CE+CPG$^3$&\colorbox{yellow}{98.41}&\colorbox{green}{86.89}&\colorbox{green}{92.80}&\colorbox{yellow}{60.27}&\colorbox{yellow}{64.15}&\colorbox{green}{67.7}&\colorbox{Turquoise}{70.86}&\colorbox{green}{79.99}&\colorbox{green}{92.64}&\colorbox{green}{66.26}&\colorbox{green}{94.57}&\colorbox{green}{82.29}&\colorbox{Turquoise}{63.56}&\colorbox{Turquoise}{95.48}&\colorbox{green}{73.26}&\colorbox{Turquoise}{88.34}&\colorbox{Turquoise}{80.27}&\colorbox{green}{65.63}&\colorbox{Turquoise}{77.66}\\
    
CE+CPG$^5$&\colorbox{Turquoise}{98.40}&\colorbox{Turquoise}{87.53}&\colorbox{Turquoise}{93.03}&55.87&61.67&\colorbox{Turquoise}{68.89}&\colorbox{yellow}{73.44}&\colorbox{yellow}{81.32}&\colorbox{Turquoise}{92.96}&\colorbox{yellow}{66.84}&\colorbox{yellow}{95.04}&\colorbox{Turquoise}{82.98}&\colorbox{green}{63.11}&\colorbox{green}{95.43}&\colorbox{yellow}{76.69}&\colorbox{green}{87.21}&\colorbox{green}{79.82}&\colorbox{yellow}{66.80}&\colorbox{yellow}{78.42}\\
CE+CPG$^7$&\colorbox{green}{98.31}&\colorbox{yellow}{87.63}&\colorbox{yellow}{93.22}&\colorbox{Turquoise}{60.14}&\colorbox{Turquoise}{64.07}&\colorbox{yellow}{70.23}&\colorbox{green}{70.23}&\colorbox{Turquoise}{81.18}&\colorbox{yellow}{93.07}&\colorbox{Turquoise}{66.34}&\colorbox{Turquoise}{94.93}&\colorbox{yellow}{84.06}&\colorbox{yellow}{66.74}&\colorbox{yellow}{95.63}&\colorbox{Turquoise}{73.29}&\colorbox{yellow}{88.97}&\colorbox{yellow}{81.39}&\colorbox{Turquoise}{66.01}&75.14\\
    
    \bottomrule
  \end{tabular}}
  \caption{Class-wise mIoU results obtained by DeepLabv3-Resnet50 on the Cityscapes validation set. Yellow indicates the method with the top mIoU value, blue represents the second highest, green signifies the third, and colorless identifies the method with the lowest.}
  \label{table1}
\end{table*}

\begin{table}
  \centering
  \resizebox{\linewidth}{!}{
  \begin{tabular}{@{}lccccccccccccc@{}}
    \toprule
    \multirow{2}{*}{CPG$^3$} & $\alpha$	&	0 &	0.2 &	0.4 &	0.6 &	0.8 &	0.9 & 1	& 1.1	& 1.2 &	1.4 &	1.6 &	2\\
	& miou(\%) &	76.75 &	78.20	& 78.22 &	78.57 &	78.73 &	78.69 &	\colorbox{yellow}{\textbf{79.00}}	& 78.10 &	78.68 &	78.79 &	78.84 &	78.12\\
		\cline{1-14}
	\multirow{2}{*}{CPG$^5$}	& $\alpha$	& 0	& 2	& 4	& 6	& 8	& 10	& 12	&14&	16&	18& &\\
	&miou(\%)&	76.75	&77.78&	78.75&	79.00&	\colorbox{yellow}{\textbf{79.23}}	&79.14	&79.05&	78.80&	79.12&	78.80& &\\
		\cline{1-14}
	\multirow{2}{*}{CPG$^7$}	&	$\alpha$	&	0	&	1	&	2	&	3	&	4	&	5	&	6	&	7		&8		&9	&	10	& \\
	&	miou(\%)		&76.75	&	78.96	&	79.20	&	79.12	&	79.45	&	\colorbox{yellow}{\textbf{79.50}}	&	79.17	&	78.93	&	79.38		&79.31	&	79.33	& \\
    \bottomrule
  \end{tabular}}
  \caption{Relationship between CPG loss weight $\alpha$ and mIoU. Different sizes of convolution kernels are used in the experiment, and the test results are based on the CityScapes validation set and the DeepLabv3-Resnet50.}
  \label{table2}
\end{table}

\subsection{Object Boundary Improvement}
In this section, we present segmentation details near the objects' borders, focusing on DeepLabv3-Resnet50 and the Cityscapes dataset.

Figure 5 displays the input image and category probabilities predicted by the networks. CPG loss enhances the visibility of black-and-white color boundaries at the edges, improving prediction accuracy, particularly for categories with intricate boundaries, as illustrated in the third row with vegetation.

Figure 6 depicts the predictive probability of contiguous pixels. Panels (a) and (b) correspond to the 1st row of Figure 5, while (c) and (d) correspond to the 2nd row, both featuring the category "pole." Poles, being slender, typically pose segmentation challenges. In our test, CPG$^3$ loss significantly boosts its validation mIoU from 63.71\% to 67.7\%. Panels (a) and (b) represent the probability results for pixels in the 78th row, while (c) and (d) show the probability of the 104th column. Noticeably steeper curves in (b) and (d) compared to (a) and (c) indicate improved edge detection, with peak and valley values closer to 1 and 0, respectively.

\subsection{Ablation Study}
In this section, we conduct tests using CPG loss with different weights ($\alpha$) and various convolution kernel sizes outlined in Equation 15. The experiments are performed on the DeepLabv3-Resnet50 network using the Cityscapes dataset, testing convolution kernel sizes of 3, 5, and 7. Results are presented in Table 2.

The experimental outcomes reveal that configuring the convolution kernel size of CPG loss to 3, 5, and 7 enhances the network's performance. The optimal results are as follows: CPG$^3$ ($\alpha$=1.0) improves mIoU by 2.252\%, CPG$^5$ ($\alpha$=8.0) improves mIoU by 2.483\%, and CPG$^7$ ($\alpha$=5.0) improves mIoU by 2.754\%. Subsequently, we employ CPG$^M$ using their respective best $\alpha$ values in the following experiments, and select visualized results are shown in Figure 7.

\begin{figure*}
  \centering
   \includegraphics[width=1.0\linewidth]{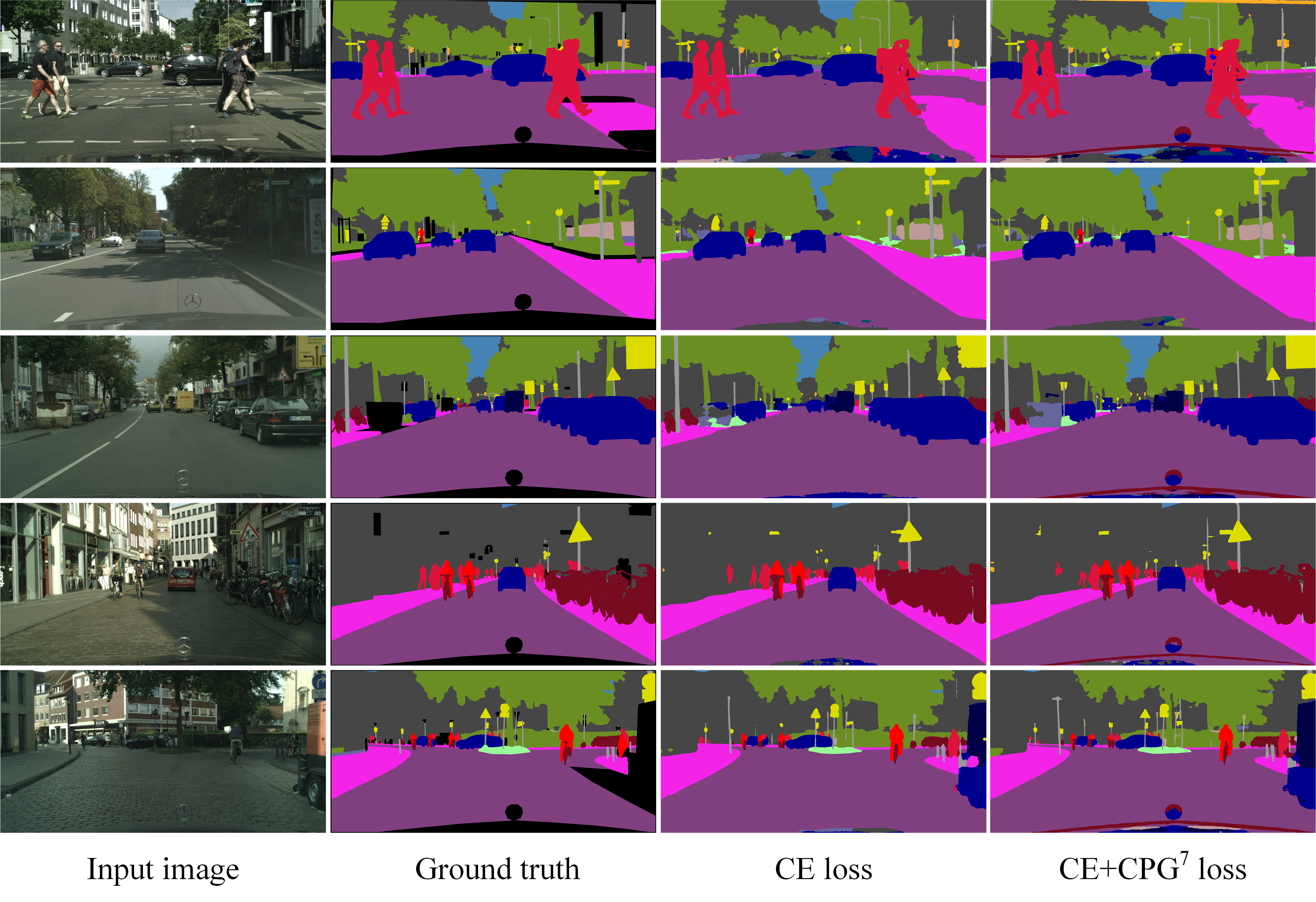}
   \caption{Selected qualitative results on the Cityscapes validation set. The tests are based on DeepLabv3-Resnet50. In the ground truth (GT), the black region signifies an undefined type. The segmentation results with CE+CPG Loss exhibit improved boundary details, particularly for lower mIoU categories such as pole, traffic sign, and traffic light.}
   \label{figure7}
\end{figure*}

In Table 1, we compare the class-wise mIoU when using CE loss alone and when adding CPG loss as an auxiliary loss (where CE loss alone corresponds to the case when $\alpha$=0 in Table 2). The mIoU of almost all categories has been significantly improved by CPG loss, with traffic light, rider, and truck increasing by more than 6\%.

\begin{table}
	\centering
		\begin{minipage}[t]{0.45\linewidth}
  \begin{adjustbox}{valign=t}
  \raggedright
  \resizebox{\linewidth}{!}{
  \begin{tabular}{@{}l|c|c|c@{}}
    \toprule
    Dataset & Net & CE(\%) &  CE+CPG(\%)\\
    \midrule
    \multirow{3}{*}{Cityscapes}& HRNetv2-OCR & 	81.56 & 	\textbf{81.90}\\
&DeepLabv3-Resnet50 & 	76.75 & 	\textbf{79.00}\\
&LRASPP\_MobileNet\_V3\_Large & 	72.26 & 	\textbf{73.76}\\
\midrule
\multirow{3}{*}{ADE20K}& HRNetv2-OCR & 	44.76	 & \textbf{47.66}\\
&DeepLabv3-Resnet50 & 	40.18 & 	\textbf{41.25}\\
&LRASPP\_MobileNet\_V3\_Large	 & 33.59	 & \textbf{35.11}\\
\midrule
\multirow{3}{*}{COCO-Stuff}& HRNetv2-OCR	 & 39.79	 & \textbf{40.71}\\
&DeepLabv3-Resnet50	 & 32.95 & 	\textbf{33.50}\\
&LRASPP\_MobileNet\_V3\_Large & 	29.25	 & \textbf{29.45} \\
    
    \bottomrule
  \end{tabular}}
  \end{adjustbox}
  \caption{mIoU results with CPG loss across various networks and standard datasets. CPG$^7$ is applied to the HRNetV2-OCR network and COCO-Stuff$\&$DeepLabv3-Resnet50, while CPG$^3$ is used in other tests.}
  \label{table3}
  
		\end{minipage}%
 \hspace{10pt}
		\begin{minipage}[t]{0.45\linewidth}
  \begin{adjustbox}{valign=t}
  \raggedleft
  \resizebox{\linewidth}{!}{
  \begin{tabular}{@{}cc|cccc@{}}
    \toprule
     \multicolumn{2}{c|}{Method}&\makecell*[c]{ mIoU\\(\%)}&\makecell*[c]{Inf. Ti-\\me (sec)}&\makecell*[c]{Backward \\Time (sec)}&\makecell*[c]{Batch Ti-\\me (sec)}\\
     \cline{1-6}
    \midrule
    \multirow{8}{*}{DeepLabv3-Resnet50+}&BCE&76.750 &0.328 &1.634 &1.962 \\

&BCE+RMI&79.875 &+0.036 &+0.033 &+0.069\\ 
&BCE+CPG$^3$&79.002 &+0.027 &+0.016 &+0.043\\ 
&BCE+CPG$^5$&79.233 &+0.039 &+0.039 &+0.078 \\
&BCE+CPG$^7$&79.504 &+0.052 &+0.050 &+0.102 \\
&BCE+RMI+CPG$^3$&80.138 &+0.067 &+0.050 &+0.117\\ 
&BCE+RMI+CPG$^5$&79.977 &+0.076 &+0.072 &+0.148 \\
&BCE+RMI+CPG$^7$&80.046 &+0.092 &+0.083 &+0.175  \\
    \bottomrule
  \end{tabular}}
  \end{adjustbox}
  \caption{Evaluation of CPG loss on Cityscapes includes MIoU testing on the validation set and time consumption testing on the training set. Each testing batch comprises 12 images distributed across 4 GPUs. We separate a complete training batch into inference and backward time. It's essential to note that inference time consists of network and loss calculation. BCE loss's time data serves as the baseline, and the time for other losses is relative to it.}
  \label{table4}
  
		\end{minipage}%
\end{table}

    

\subsection{Quantitative Evaluation}
In this section, we examine the impact of CPG loss training on three networks using the same dataset, as presented in Table 3 with detailed results. The experiments demonstrate that the application of CPG loss enhances the performance in each group, affirming its suitability for various networks and datasets.

In the HRNetV2-OCR experiment, we observed that CPG$^3$ did not contribute to the improvement of the training result, while CPG$^7$ yielded better results. We speculate that this discrepancy arises from the direct output of HRNetV2-OCR being 1/4 the size of the original image, and the final output is obtained through bilinear interpolation of the direct output result. This may affect the performance of small convolution kernels in CPG loss.

\subsection{Comparision with Region Mutual Information Loss}
We experimentally compare CPG with Region Mutual Information (RMI)\cite{NEURIPS2019_a67c8c9a} loss. Additionally, we assess the potential synergy of these two loss functions.

RMI loss aims to model pixel relationships in an image for improved segmentation. Pixels and their neighbors collectively represent a pixel, creating a multi-dimensional point based on a small image region. This approach transforms the image into a multi-dimensional distribution of these high-dimensional points. RMI loss maximizes mutual information (MI) between predictions and ground truth. We utilize the official implementation\footnote{https://github.com/ZJULearning/RMI} with default parameters.



In our tests, BCE loss serves as the primary loss with a weight of 1.0. RMI loss has a weight of 0.5, and CPG losses use the optimal weights from Table 2. Evaluation results in Table 4 reveal that the time consumption of CPG losses increases with their convolution kernel size, and only CPG$^3$ takes less time than RMI loss. By moving the generation of $Grad^{GT,Boundary}$ from loss computation to data loading, this time could be further reduced. The performance of CPG loss is not as good as RMI loss, but their gap is only 0.37\%. However, it also demonstrates that combining CPG and RMI loss yields better results than using either of them alone, indicating their compatibility.

\section{Conclusion}
This paper introduces CPG loss as a method to enhance semantic segmentation network performance. CPG loss utilizes neighboring pixels to compute the probability gradient of a pixel, establishing pixel relationships in the image. By maximizing the similarity of the probability gradient between prediction and ground truth, the predicted results align closely with the ground truth. To focus on object edges, CPG loss specifically calculates the error of the gradient at the boundary. CPG loss is intuitive, easy to implement, and applies with minimal memory usage during training without altering network structure. Tests demonstrate consistent performance improvements across different datasets, and compatibility with RMI loss.

\bibliographystyle{unsrtnat}


\end{document}